\pdfoutput=1
\documentclass[11pt]{article}

\usepackage[preprint]{acl}

\usepackage{multirow}
\usepackage{adjustbox} % for adjusting the table size
\usepackage{setspace} % for item spacing
\usepackage{times}
\usepackage{pdfpages}
\usepackage{latexsym}
\usepackage{float}
\usepackage{wrapfig}
\usepackage{amsmath}
\usepackage{authblk}
\usepackage{subcaption}
\usepackage{graphicx}
\usepackage{adjustbox}
\usepackage{url}
\usepackage[T1]{fontenc}
\usepackage{enumitem}
\usepackage[utf8]{inputenc}
\usepackage{lineno}
\usepackage[prependcaption,textsize=tiny]{todonotes}
\definecolor{ao}{rgb}{0.0, 0.5, 0.0}

\usepackage{inconsolata}

\title{CSSL: Contrastive Self-Supervised Learning for Dependency Parsing on Relatively Free-Word-Ordered and  Morphologically-Rich Low-Resource Languages}

\author{\textbf{Pretam Ray}\textsuperscript{$\star$},\textbf{ Jivnesh Sandhan}\textsuperscript{$\dagger$},\textbf{ Amrith Krishna}\textsuperscript{$\diamond$} \textbf{and} \textbf{ Pawan Goyal}\textsuperscript{$\star$}
\vspace{-0.7em}
  \\
  \textsuperscript{$\star$} IIT Kharagpur,  \textsuperscript{$\dagger$} IIT Dharwad, 
  \textsuperscript{$\diamond$} Independent Researcher \\
  \texttt{pretam.ray@kgpian.iitkgp.ac.in, jivnesh@iitdh.ac.in, krishanmrith12@gmail.com}\\
\texttt{pawang@cse.iitkgp.ac.in}
  }

\begin{document}
\maketitle
\begin{abstract}
% \ak{1) Abstract doesnt convey anything informational, for the first 10 lines (anyone would ose interest by then)
% 2) Abstract doesnt align with the overall narrative}
% Significant advancements have been made in the domain of dependency parsing, with researchers introducing novel architectures to enhance parsing performance. However, the majority of these architectures have been evaluated predominantly in languages with a fixed word order, such as English. Consequently, little attention has been devoted to exploring the robustness of these architectures in the context of relatively free word-ordered languages. In this work, we examine the robustness of graph-based parsing architectures on 7 relatively free word order languages. We focus on investigating essential modifications such as data augmentation and the removal of position encoding required to adapt these architectures accordingly.
% To this end, we propose a contrastive self-supervised learning method to make the model robust to word order variations. Furthermore, our proposed modification demonstrates a substantial average gain of 3.03/2.95  points in 7 relatively free word order languages, as measured by the Unlabelled/Labelled Attachment Score metric when compared to the best performing baseline.
Neural dependency parsing has achieved remarkable performance for Low-Resource Morphologically-Rich languages. It has also been well-studied that Morphologically-Rich languages exhibit relatively free-word-order. This prompts a fundamental investigation: \textit{Is there a way to enhance dependency parsing performance, making the model robust to word order variations utilizing the relatively free-word-order nature of Morphologically-Rich languages?} In this work, we examine the robustness of graph-based parsing architectures on 7 relatively free-word-order languages. We focus on scrutinizing essential modifications such as data augmentation and the removal of position encoding required to adapt these architectures accordingly. To this end, we propose a contrastive self-supervised learning method to make the model robust to word order variations. Furthermore, our proposed modification demonstrates a substantial average gain of 3.03/2.95  points in 7 relatively free-word-order languages, as measured by the UAS/LAS Score metric when compared to the best performing baseline. \footnote{The code is available at \url{https://github.com/raypretam/cssl-DP}.}

\end{abstract}

\section{Introduction}
Dependency parsing for low-resource languages has greatly benefited from diverse data-driven strategies, including data augmentation \cite{sahin-steedman-2018-data}, multi-task learning \cite{nguyen-verspoor-2018-improved}, cross-lingual transfer \cite{10.1145/3383772}, self-training \cite{rotman-reichart-2019-deep, clark-etal-2018-semi} and pre-training \cite{sandhan2021little}. Further, incorporating morphological knowledge substantially improves the parsing performance for low-resource Morphologically-Rich languages \citep[MRLs;][]{dehdari-etal-2011-morphological, vania-etal-2018-character, dehouck-denis-2018-framework, krishna-etal-2020-keep, roy-etal-2022-meta,10.1145/3555340}. 

MRLs tend to have sentences that follow a relatively free-word-order \cite{futrell-etal-2015-quantifying, krishna-etal-2020-graph}, as structural information is often encoded using morphological markers rather than word order. In MRLs, a sentence may have different acceptable word order configurations, that preserve the semantic and structural information. However, the permutation invariance is often not reflected in their corresponding semantic space representations encoded using a pretrained model. Pretrained models typically include a position encoding component, often shown to be beneficial for tasks in languages that follow a fixed word order. However, removing the position encoding of the encoder during fine-tuning is demonstrated to be counterproductive \cite{krishna-etal-2019-poetry, ghosh2024morphologybased}.

Languages, including MRLs, tend to follow a preferred word order typology. However, such preferences are often followed for the efficiency of communication, from a cognitive, psycho-linguistic, and information-theoretic standpoint and not due to any limitations of the morphology \cite{krishna-etal-2019-poetry, clark2023crosslinguistic, xu-futrell-2024-syntactic}. For instance,  Sanskrit, a classical language \citep{coulson1976sanskrit}, predominantly consists of sentences written as verses in its pre-classic and classic literature. The majority of the available corpora in Sanskrit are written in verse form \citep{dcs}. Here, verbal cognition often takes a backseat as words are often reordered to satisfy metrical constraints in prosody \citep[\S 2]{krishna-etal-2020-graph}. Hence, these sentences appear to be arbitrarily ordered based on syntactic analysis \citep{kulkarni2015free}. In this work, we propose a self-supervised contrastive learning framework, primarily for Sanskrit, that makes the model agnostic to the word order variations within a sentence.

\begin{figure*}[htbp]
    \centering
    \begin{subfigure}{0.75\linewidth}
        \centering
        \adjustbox{width=\linewidth, keepaspectratio}{%
            \includegraphics[page=1]{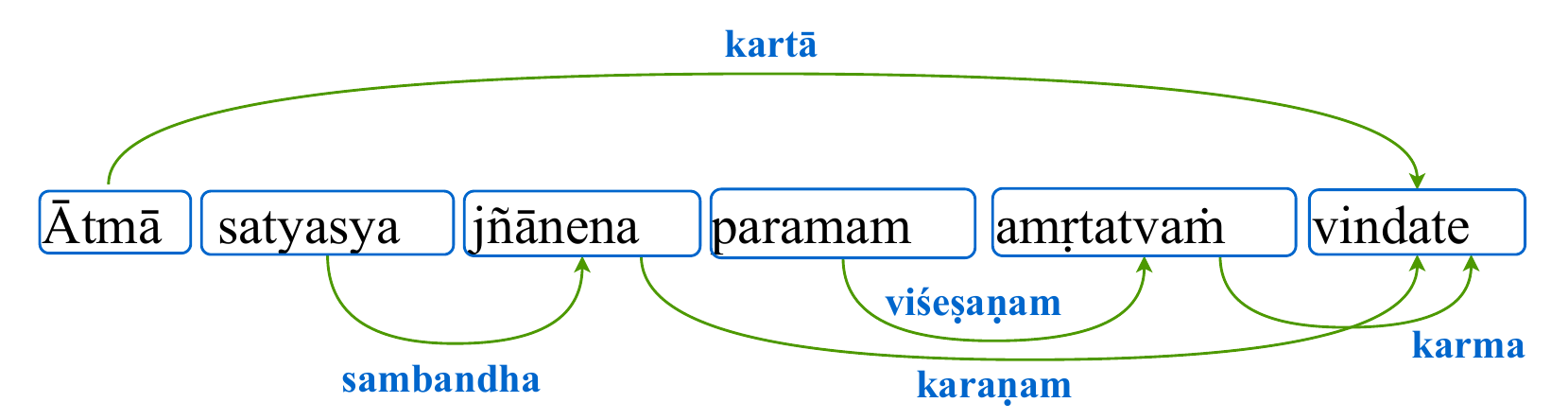}%
        }
        \caption{Original Word Order}
        \label{fig:dependency_original}
    \end{subfigure}
    \hfill
    \begin{subfigure}{0.75\linewidth}
        \centering
        \adjustbox{width=\linewidth, keepaspectratio}{%
            \includegraphics[page=1]{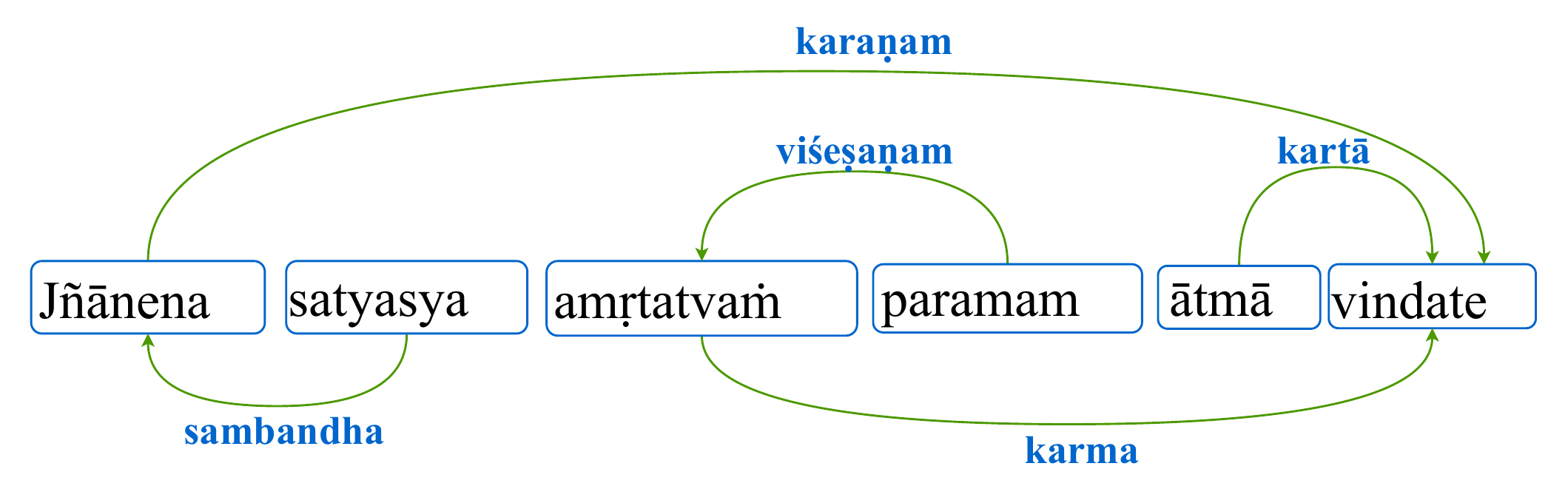}%
        }
        \caption{Permuted Word Order}
        \label{fig:dependency_permuted}
    \end{subfigure}
    \caption{The Dependency Parse of a Sanskrit sentence and its random permutation is same, exhibiting free-word-order nature of the language. Translation: "The soul attains supreme immortality through the knowledge of truth."}
    \label{fig:dependency_main}
\end{figure*}

Our Contrastive Self-Supervised Learning (CSSL) framework builds upon the recent success of using annotated pairs in contrastive learning \citet{khosla2021supervised, yue2021contrastive} to make the model permutation invariant to the arbitrary word order variations in Sanskrit \citep{Wright_1968}. Given the comprehensive morphological marking system, the core semantic essence of a sentence remains unaltered, rendering the permuted counterpart as a suitable positive pairing for contrastive learning. Figure \ref{fig:dependency_main} shows that the original sentence and its permuted counterpart having the same dependency structure. This simple use of word permutations in a sentence as positive pairs achieves substantial improvement over prior methods. Our approach, to the best of our knowledge, is the first to use a contrastive learning approach for dependency parsing.
Our proposed approach is modular and agnostic, allowing for seamless integration with any encoder architecture without necessitating alterations to the pretaining decisions. Moreover, our objective is to leverage recent advancements in parsing literature by augmenting with the CSSL framework, which would make these models more robust to word order variations. In this work, we start by examining the robustness of graph-based parsing architectures \citep{ji-etal-2019-graph,mohammadshahi-henderson-2020-graph,mohammadshahi-henderson-2021-recursive}. We believe, graph-based parsing architectures could be a natural choice to model flexible word order.  We then focus on investigating essential modifications such as data augmentation \citet{sahin-steedman-2018-data} and the removal of position encoding \citet{ghosh2024morphologybased} required to adapt these architectures accordingly. We finally show the efficacy of our approach on the best baseline \citet[RNGTr]{mohammadshahi-henderson-2021-recursive} model by integrating CSSL with it and report an average performance gain of 3.03/2.95 points (UAS/LAS) improvement over 7 MRLs. 

%Existing multi-lingual pretraining opts for default position encoding, however, this decision may not be optimal for low-resource relatively free word order MRLs. Further, building word order agnostic encoder from scratch is not feasible due to data sparsity. Our results suggest that if we simply drop position encoding in the encoder, then a mismatch in pretraining and task setting leads to suboptimal performance.

 % These strategies include the use of graph neural network-based architecture(\citeauthor{ji-etal-2019-graph}, \citeyear{ji-etal-2019-graph}), graph transformer based architecture \cite{mohammadshahi-henderson-2020-graph, mohammadshahi-henderson-2021-recursive} and data augmentation \cite{sahin-steedman-2018-data}. However, the research community has paid little consideration to morphologically rich low-resource languages that have relatively free word order. 
% The Transformer encoder or decoder is pretrained on a large-scale text corpus by completing self-supervised tasks like creating future tokens \cite{Radford2018ImprovingLU} and predicting masked tokens \cite{devlin2019bert}. The targets to be predicted in these studies are primarily word-level. Consequently, there's a chance that the sentence-level global semantics which is crucial for a morphologically rich language may not be fully captured.\textcolor{red}{need citation}.\\

Our main contributions are as follows:

\begin{itemize}[]
    \item We propose a novel contrastive self-supervised learning (CSSL) module to make dependency parsing robust for free-word-order languages.
    \item Empirical evaluations of CSSL module affirm its efficacy for 7 free word-ordered languages 
    \item We demonstrate statistically significant improvements with an average gain of  3.03/2.95  points over the best baseline on 7 MRLs. 
\end{itemize}
% Our exhaustive experimentation empirically establishes the effective ability of graph neural networks and graph transformers in the Poetry Domain. It demonstrates significant improvements with an average absolute gain of 10.36/10.47 points Unlabelled/Labelled Attachment Score (UAS/LAS) over strong baseline (\citeauthor{DBLP:conf/iclr/DozatM17}, \citeyear{DBLP:conf/iclr/DozatM17}) without data augmentation. Notably, our approach C-RNGTR surpasses our strongest baseline RNGTR, showcasing substantial improvements. C-RNGTR outperforms our strongest baseline by  14.13 (UAS) and 14.28 (LAS) points. Finally, we demonstrate the method's effective use in the Poetry domain of a genuinely low-resource language.

\section{Contrastive Self-Supervised Learning}

\begin{figure}[ht]
    \centering
    \includegraphics[width=0.8\linewidth]{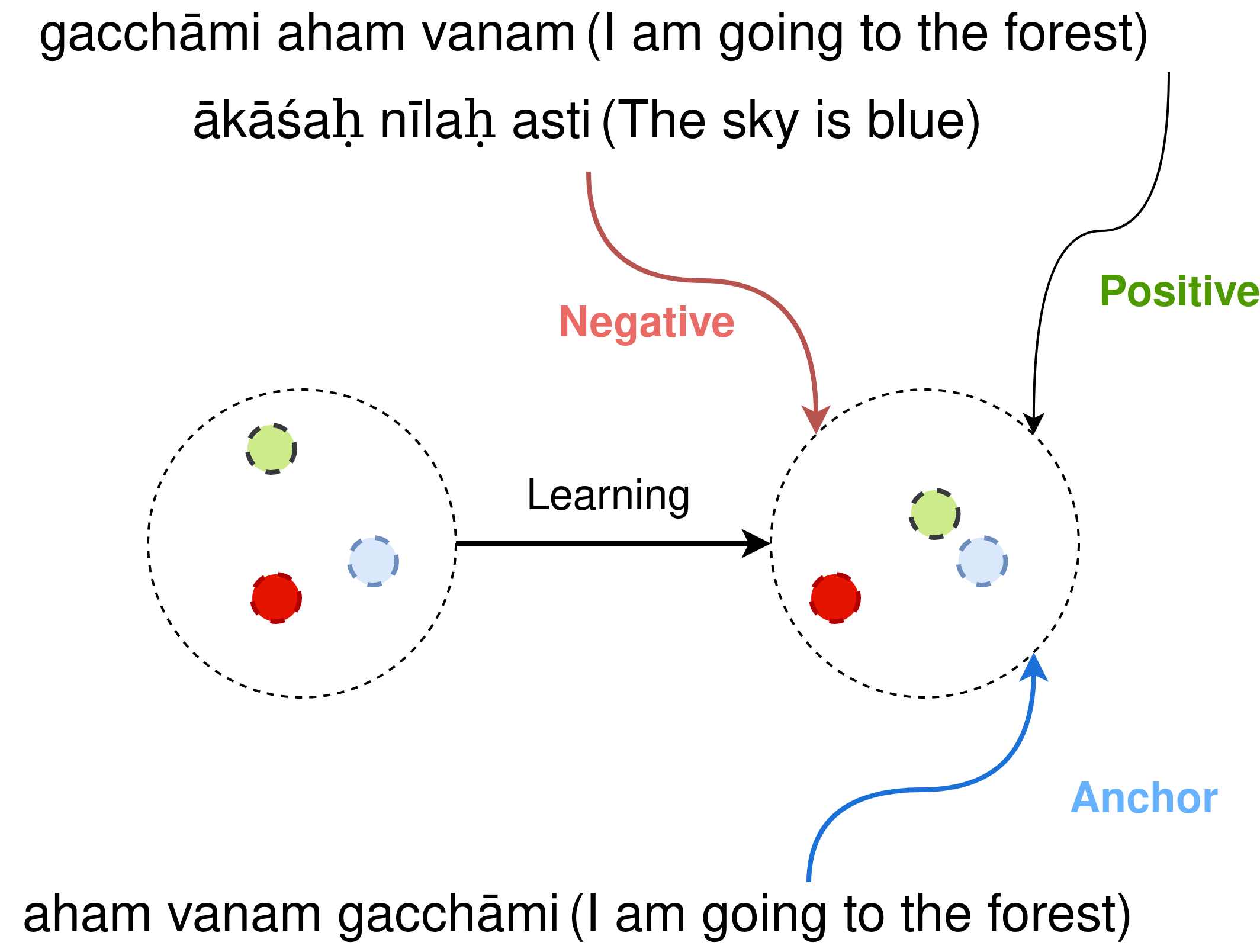}
    \caption{ The Contrastive Loss minimizes the distance between an anchor (blue) and a positive (green), both of which have a similar meaning, and maximizes the distance between the anchor and a negative (red) of a different meaning.}
    \label{fig:1}
\end{figure}

CSSL enables joint learning of representation, via contrastive learning, with the standard classification loss for dependency parsing. Here, via CSSL, we identify sentences which are word-level permutations of each other as similar sentences, and others as dissimilar sentences. The similar sentences are brought closer while  pushing dissimilar examples apart
 \cite{oord2019representation, tian2020contrastive}.
% In self supervised contrastive learning, for a given input, one needs to find positive samples, whose embedding level similarity with the input needs to be increased, and negative samples, whose embedding similarity with the original input, needs to be decreased. 
As shown in Figure \ref{fig:1}, the original sentence serves as an anchor point, while its permutations represent positive examples, juxtaposed with randomly generated sentences serving as negative examples. 
For a given input, when selecting a dissimilar sample, we choose a random sentence that clearly differs significantly from any permutation of the given sentence.
%This section outlines the proposed CSSL module. MRLs depend less on strict word order, thus, this module enhances the encoder's resilience to accommodate variations in word order.CSSL facilitates the acquisition of robust representations by bringing similar examples closer together and.
% In languages with flexible word order, intricate morphology allows for relaxed constraints on word order. This means that rearrangements of words can maintain semantic equivalence, particularly when weak projectivity is involved. Morphologically Rich Languages, with their detailed marking systems, preserve the fundamental semantic meaning of sentences even when word order is changed. Thus, these rearranged versions serve as suitable counterparts for contrastive learning.

%We induce the similarity based on permutation of word order (the set of words remains the same, just order changes), and not semantic similarity.  Even if a randomly selected sentence happens to be semantically similar, our training objective still favors the shuffled word order counterpart.

 % In this work, we investigate sentences written as word order variations of one another, resulting in a higher semantic affinity. As a result, it is essential that the representation of a certain sentence is consistent across both original and permuted samples. In , connections are made of the contrastive loss to maximization of mutual information between different views of the data. \\
\begin{figure}[ht]
    \includegraphics[width=1\linewidth]{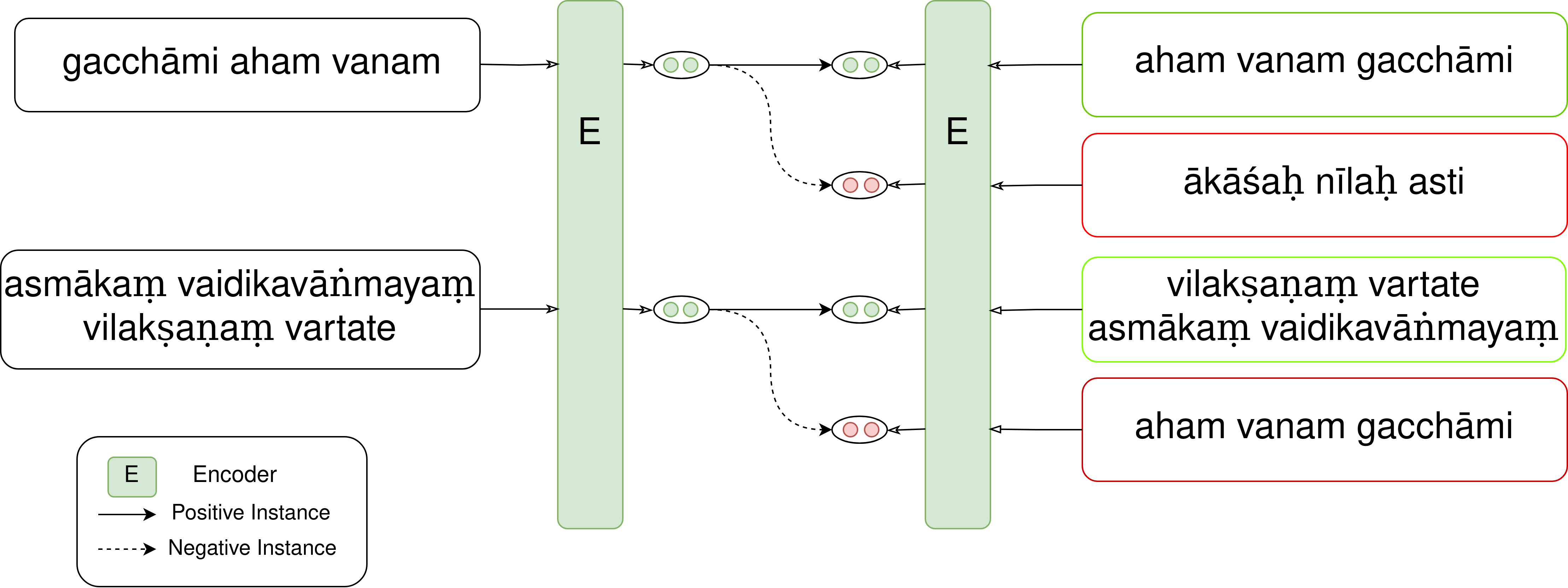}
    \caption{Schematic illustration of the proposed approach for Sanskrit. Self-supervised CSSL leverages the sentence and its permutation pairs as positives and other in-batch instances as negatives.}
    \label{fig:2}
\end{figure}
% \ak{A lot of it seems to be redundant with the introduction. Ideally hte intro needs to be rewritten
% }
Formally, as shown in Figure \ref{fig:2} for a sentence \(X_i\) (anchor example), its representation should be similar to the permuted instance \(X_i^+\) as permutation \footnote{Refer to Appendix\ref{permutation} for the algorithm to generate the permutations. } does not alter the meaning of a sentence in Sanskrit. However, the representation will differ from a random sentence \(X_i^-\) (negative example). Therefore, the distance between the appropriate representations of \(X_i\) and \(X_i^+\) is expected to be small. Thus, we can develop a contrastive objective by considering (\(X_i\), \(X_i^+\)) a positive pair and $N-1$ negative pairs (\(X_i\), \(X_i^-\)) :
\begin{multline*} \label{eq1}
    \mathcal{L}_{\text {CSSL}} = - \log \frac{\exp \left(\boldsymbol{z}_i \cdot \boldsymbol{z}_{i^+} / \tau\right)}{\sum_{a \in {N}} \exp \left(\boldsymbol{z}_i \cdot \boldsymbol{z}_a / \tau\right)}
\end{multline*}
where $N$ represents a batch, $z_i$ represents the representation vector of the anchor sample, $z_i^+$ denotes the representation vector for the positive sample (permuted sample), $z_a$ represents the representation vector for a sample in the batch ($N$ different samples), and $\tau$ is a temperature parameter that controls the concentration of the distribution. 
For all representation vectors, we employ pooled sentence embedding \cite{reimers-2020-multilingual-sentence-bert} for the CSSL loss. 
Therefore, our final loss is: 
\begin{equation}
\mathcal{L} = \mathcal{L}_{\text {CSSL}} + \mathcal{L}_{\text {CE}}
\end{equation}
% where $L_{cts}$ is the contrastive loss and $L_{ce}$ is the cross-entropy loss based on the ground truth arcs and head relations of the sample $X_i$.
 The classification loss $L_{CE}$ is the cross-entropy loss applied only to token-level labels of the original training input. The scorer is based on biaffine-scorer \cite{dozat2017deep}, which tries to independently maximize the local probability of the correct head word for each word. 
 % \pg{You are still not providing details. How is $z_a$ computed? What is $L_{CE}$? How is it computed? If it is different for different encoder architectures, please mention, and provide details in the appendix. Cross-entropy details}
 % \ak{agreed with the above comment}
% During training, we permute the sentence's word order to make it a positive example, while the remaining sentences in the batch are treated as negative examples for contrastive learning. 

\section{Experiment}  
% In this section, we evaluate our framework against \newcite[RNGTr]{mohammadshahi-henderson-2021-recursive}. We also show that our framework consistently outperforms the rotation-based DA technique \newcite[crop-rotate]{sahin-steedman-2018-data} on 7 MRLs for dependency parsing. 
% We carefully study the regularization effects of our framework and empirically demonstrate that it leads to a more dispersed representation space with a semantic structure better encoded.
 
% \begin{table}[h]
%     \centering
%     \begin{tabular}{l|l|l|l}
%     \hline
%         \multicolumn{1}{l|}{\textbf{Treebank}}  &\multicolumn{1}{c|}{\textbf{train}} & \multicolumn{1}{c|}{\textbf{dev}} &\multicolumn{1}{c}{\textbf{test}}  \\ \hline
%         Sanskrit-STBC & 2800 & 300 & 300 \\ \hline
%         UD-Turkish\_IMST & 3435 & 1100 & 1100 \\ \hline
%         UD-Gothic\_Proeil & 3387 & 985 & 1029 \\ \hline
%         UD-Telugu\_MTG & 1051 & 131 & 146 \\ \hline
%         UD-Hungarian\_Szeged & 910 & 441 & 449 \\ \hline
%         UD-Ancient\_Hebrew\_PTNK & 730 & 439 & 410 \\ \hline
%         UD-Lithuanian\_ALKSNIS & 2341 & 617 & 684 \\ \hline
%         UD-English\_EWT & 12544 & 2001 & 2077 \\ \hline
%     \end{tabular}
%     \caption{Treebank Statistic. train,dev and test mean the number of sentences in each set.}
%     \label{tab:stats}
% \end{table}
\subsection{Dataset and metric}
% \ak{As far as the narrative is concerned, I would be surprised what is the need of a primary dataset and additional datasets. So far, there is no mention of Sanskrit, or mention of any specific factor that is characteristic (or even unique) to Sanskrit. Again, please go through the reviews. The reviewers were confused at the same point. My opinion is that we should make it a Sanskrit centric paper right from the intro. Make our case using the properties of Sanskrit. Then in the experiments we should say, "Hey our propsoed approach generalizes to numerous morphologically rich languages" }
As our primary benchmark dataset, we 
utilize the Sanskrit Treebank Corpus \cite[][STBC]{kulkarni-2013-deterministic}. From STBC, we use a train and dev split of 2,800 and 1,000 respectively. Further, we employ a test set comprising 300 sentences, drawn from the classical Sanskrit work, \textit{Śiśupāla-vadha} \cite{ryali2016}. 

Moreover, from Universal Dependencies \cite[][UD-2.13]{de-Marneffe}, we choose 6  additional Morphologically-Rich low-resource languages, namely, Turkish, Telugu, Gothic, Hungarian, Ancient Hebrew, and Lithuanian.\footnote{The statistics of each of the treebanks used for our experiments is mentioned in Table \ref{tab:stats} in the Appendix.} Please note that all the seven languages are chosen from diverse language families and are typologically diverse. Our experiments are primarily focused on a low-resource setting. However, we also show how the framework performs on high-resource MRL. We also experiment with English which is a fixed-ordered high-resource language. Here, we use a training set of 12,544 sentences. We use standard UAS/LAS metrics \cite{mcdonald-nivre-2011-analyzing} for evaluation.

%\paragraph{Language Selection Criteria:}We choose 7 low-resource languages from 7 typological families (Table \ref{tab:stats}). We restrict morphologically rich languages from various language families, script families, morphological complexity, and domains.

% \subsection{Evalutation Metric}
% We report the performance of each system using the  Unlabelled Attachment Score (UAS) and the Labelled Attachment Score (LAS) as metrics. UAS is the fraction of correct tokens with correct predicted heads and LAS is the fraction of tokens with thecorrect label \cite{mcdonald-nivre-2011-analyzing}.
\begin{table}[ht]
    \centering
    \begin{tabular}{l|cc}
    \multicolumn{1}{c|}{Model} & UAS                  & LAS                  \\ \hline \hline
    G2GTr (Transition-based) & 85.75              & 82.21             \\
    GNN (Graph-based)        & 88.01              & 82.8              \\
    RNGTr (Graph-based)  & 89.62 & 87.43  \\\hline \hline
    RNGTr (NoPos)       & 80.78                & 78.37                \\
    RNGTr (DA)   & 90.38 & 88.46 \\
    Prop. System (CSSL)         & \textbf{91.86}       & \textbf{89.38} \\ \hline \hline
    CSSL + DA & 92.43 & 90.18 \\ \hline
    \end{tabular}
    \caption{Comparison of graph-based parsers on Sanskrit STBC dataset. We modify the best baseline RNGTr by integrating the proposed method (CSSL) to compare against variants, removing position encoding (NoPos) and data augmentation (DA).
    The best performances are bold-faced. The results (CSSL vs DA) and (CSSL vs DA+CSSL) are statistically significant as per the t-test with a p-value < 0.01 for the LAS metric.}
    \label{tab:Methods on Sanskrit}
\end{table}
\begin{table*}[ht]
% \scalebox{0.9}{
    \centering
    \begin{tabular}{l|l|cc|cc|ccc}
    \hline
        \multicolumn{1}{l}{} & ~ & \multicolumn{2}{c|}{RNGTr} & \multicolumn{2}{c|}{RNGTr + DA} &\multicolumn{2}{c}{RNGTr + CSSL}  &  \\ \cline{3-9}
        
        \multicolumn{1}{l}{Language} & \multicolumn{1}{l}{Setting}& \multicolumn{1}{|p{1.5cm}}{\hspace{0.32cm}UAS} &  \multicolumn{1}{c|}{LAS} & \multicolumn{1}{p{1.5cm}}{\hspace{0.32cm}UAS} &  \multicolumn{1}{c|}{LAS} & \multicolumn{1}{p{1.4cm}}{\hspace{0.32cm}UAS} &  \multicolumn{1}{c}{LAS} \\ \hline
        % \multicolumn{1}{l|}{Sanskrit-STBC} & 89.62 &	87.43 & 90.38 & 88.46 & \textbf{91.86} & \textbf{88.89} \\ \hline
        \multicolumn{1}{l|}{Turkish-IMST} & LRL & 72.86 & 71.99 & 74.18 & 72.96 & \textbf{78.21} & \textbf{74.69} \\ \hline
        \multicolumn{1}{l|}{Telugu-MTG} & LRL & 90.02	& 80.34 & 91.86 & 81.51 & \textbf{93.79} & \textbf{85.67} \\ \hline
        \multicolumn{1}{l|}{Gothic-POIEL}  & LRL & 86.59 & 81.28 & 88.61 & 82.93 & \textbf{89.15} & \textbf{84.19} \\ \hline
        \multicolumn{1}{l|}{Hungarian-SZEGED} & LRL & 88.13 	& 84.93 &	90.02  & 86.65 & \textbf{91.65} &\textbf{87.28} \\
        \hline
        % \multicolumn{1}{l|}{Welsh} & ~ 	& ~ &	~  & ~ & ~ & ~ \\
        % \hline
        \multicolumn{1}{l|}{Ancient Hebrew-PTNK} & LRL & 90.76 & 86.42 & 91.43  & 87.12 & \textbf{92.35} & \textbf{88.68} \\ \hline
        \multicolumn{1}{l|}{Lithuanian-ALKSNIS} & LRL & 87.63 &	83.27  & 88.41	& 84.79 & \textbf{89.82} & \textbf{86.45} \\ \hline \hline
        \multicolumn{1}{l|}{Turkish-PENN} & HRL & 82.31 & 76.23 & 85.57 & 78.19 & \textbf{88.43} & \textbf{80.82} \\ \hline \hline
        \multicolumn{1}{l|}{English-EWT} & non-MRL & \textit{92.08} & \textit{90.23} & \textbf{\textit{93.76}} & \textbf{\textit{92.16}} & \textit{93.19} & \textit{90.71} \\ 
        
        \hline
    \end{tabular}
    \caption{Performance comparison on the RNGTr model on UD Treebanks, RNGTr + DA (Data Augmentation) and RNGTr + CSSL module. The best performances are bold-faced. Our results (CSSL) are statistically significant compared to both RNGTr and RNGTr + DA for each language as per the t-test with a p-value < 0.01 for the LAS metric. LRL stands for low-resource MRL, HRL means high-resource MRL.}
    \label{tab:contrastive}
\end{table*}
\paragraph{Baselines:}
% We utilize \citeauthor{mohammadshahi-henderson-2021-recursive} (\citeyear{mohammadshahi-henderson-2021-recursive}, \textbf{RNGTr}) architecture which iteratively refines arbitrary graphs through recursive operations that use pre-trained mBERT which is trained using only cross entropy based loss function. \textcolor{red}{change needed}

We utilize \citeauthor{mohammadshahi-henderson-2020-graph} (\citeyear{mohammadshahi-henderson-2020-graph}, \textbf{G2GTr}), a transition-based dependency parser. 
% Subsequently, we assess the performance of four graph-based dependency parsing approaches. 
% Firstly, we consider the \citeauthor{DBLP:conf/iclr/DozatM17} (\citeyear{DBLP:conf/iclr/DozatM17}, \textbf{BIAFF}) approach, which employs a graph-based framework with a BIAFFINE attention mechanism. We also examine \citeauthor{nguyen-etal-2021-trankit} (\citeyear{nguyen-etal-2021-trankit}, \textbf{TRANKIT}) which extends the BIAFFINE model to support multilingualism with \citeauthor{conneau2020unsupervised} (\textbf{XLM-R}) as base Transformer. 
Furthermore, we explore \citeauthor{ji-etal-2019-graph} (\citeyear{ji-etal-2019-graph}, \textbf{GNN}) a graph neural network-based model that captures higher-order relations in dependency trees. Finally, we examine Graph-to-Graph Non-Autoregressive Transformer proposed by \citeauthor{mohammadshahi-henderson-2021-recursive} (\citeyear{mohammadshahi-henderson-2021-recursive}, \textbf{RNGTr}) which iteratively refines arbitrary graphs through recursive operations.
% \subsection{Experimental Setup}
% The encoder is initialized with a pre-trained multilingual BERT \cite{devlin2019bert} model with 12 self-attention layers from Huggingface. All hyper-parameters are provided in Appendix A.\\
% Following the work by \cite{mohammadshahi-henderson-2021-recursive} we employ the BERT wordpiece tokenizer on each corpus word and encode the resulting sub-words. Dependency ties between two words are specified as a relationship between their first sub-words. We create a new relationship between non-first sub-words and their corresponding first sub-word as the head. The decoder simply uses the encoder embedding of each word's initial sub-word to predict dependency relations.\\

\paragraph{Hyper-parameters:} 
% We implement our CSSL module in RNGTr architecture which uses a pre-trained mBERT model (110M parameters) from Huggingface transformers \cite{wolf-etal-2020-transformers}. 
For RNGTr model, we use the same architecture from the work of \newcite{mohammadshahi-henderson-2020-graph} which uses pre-trained mBERT \cite{wolf-etal-2020-transformers} as the encoder and an MLP and biaffine followed by softmax for the decoder. We adopt the RNGTr codebase with hyperparameter settings as follows: the batch size is 16,  the learning rate as 2e-5, the number of transformer blocks as 12 and for the decoder 2 Feed Forward Layers, and the remaining hyperparameters are the same.

\subsection{Results}
In Table \ref{tab:Methods on Sanskrit}, we benchmark graph-based parsers on the Sanskrit STBC dataset.  Our proposed contrastive loss module is standalone and could be integrated with any parser.\footnote{Refer to Appendix \ref{modular nature} for empirical evidence.} Thus, we modify the best baseline RNGTr by integrating the proposed method (CSSL) and comparing it against variants, removing position encoding (NoPos), and augmenting data augmentation (DA) \footnote{Refer to Appendix \ref{DA} for the algorithm used in Data Augmentation.}. Table \ref{tab:Methods on Sanskrit} illustrates that the proposed framework adds a complementary signal making robust word order representations to RNGTr by improving 2.24/1.95 points in UAS/LAS scores. The performance significantly drops (8.8/9.0 UAS/LAS) when position embeddings are removed (vs. Pos kept) from RNGTr due to train-test mismatch in pretraining and fine-tuning steps.
 Moreover, our method outperforms data augmentation technique (DA) \cite{sahin-steedman-2018-data} by 1.48/0.92 points (UAS/LAS) when integrated with the RNGTr baseline. We integrate CSSL on top of an RNGTr+DA system and observe statistically significant improvements of 0.57/0.80 points (UAS/LAS), suggesting the proposed method complements the data-augmentation technique. 

\paragraph{Results on multilingual experiments:}
In this section, we investigate the efficacy of CSSL module in multi-lingual settings. For all MRLs, the trend is similar to what is observed for Sanskrit. Table \ref{tab:contrastive} reports results on 6  other Morphologically-Rich languages in low-resource settings.  Our approach averages 3.16/3.12 higher UAS/LAS scores than the usual cross-entropy-based RNGTr baseline. Our system outperforms the rotation-based DA technique with an average increase of 1.74/1.83 in UAS/LAS scores. Here, as expected, our proposed CSSL approach outperforms the standard RNGTr and DA approaches for all the languages, except English. English is not an MRL and it relies heavily on configurational information of the words to understand sentence structure. The DA approach performs better by 0.57/1.45 UAS/LAS scores than our framework. However, it is interesting to note that CSSL still outperforms the RNGTr baseline by 1.11/0.48 UAS/LAS, possibly due to robustness of permutation invariant representation learning we employ in CSSL. \\
As illustrated in Table \ref{tab:Methods on Sanskrit}, it is evident that combining CSSL with DA surpasses CSSL alone by approximately 0.5 points, exhibiting a 2-point enhancement over DA.\\
We also experiment with Turkish on UD\_Turkish-PENN Treebank in a high-resource setting, having 14,850 sentences in the training set. Our CSSL framework outperforms usual cross-entropy technique by 6.12/4.59 in UAS/LAS scores and outperforms the DA technique by 2.96/2.63 in UAS/LAS scores. The significant increase in score can be attributed to the greater number of training examples. 

\section{Conclusion}
In this work, we investigated the robustness of graph-based parsing architectures across 7 languages characterized by relatively flexible word order. 
%MRLs rely less on strict word order and instead utilize morphological markers to encode sentence structure.
We introduced a self-supervised contrastive learning module aimed at making encoders insensitive to variations in word order within sentences. 
Additionally, the modular nature of our approach enables seamless integration with any encoder architecture without necessitating modifications to pretraining decisions.
To the best of our knowledge, our approach represents the first utilization of contrastive learning techniques for dependency parsing to address challenges arising from variable word order in low-resource settings. Finally, we demonstrate the effectiveness of our approach by integrating it with the RNGTr architecture \newcite{mohammadshahi-henderson-2021-recursive}, reporting an average performance improvement of 3.03/2.95 points (UAS/LAS) across the 7 MRLs.
% While significant strides have been made in dependency parsing for languages with fixed word order like English, our research illuminates a critical gap in evaluating these architectures across more diverse linguistic structures. 

% Our findings demonstrate the efficacy of the proposed modifications, particularly the incorporation of a contrastive loss objective, in enhancing parsing performance across the languages under scrutiny.
% \\
% Future work could consider extending this method for dependency parsing in poetry data, where more intricate word orderliness is found. Future research in this domain could further refine and generalize these modifications to encompass a broader spectrum of languages, ultimately advancing the field of dependency parsing in linguistically diverse contexts.

\paragraph{Limitations}
We could not evaluate on complete UD due to limited available compute resources (single GPU); hence, we selected 7 representative languages for our experiments.
\paragraph{Ethics Statement}
We do not foresee any ethical concerns with the work presented in this manuscript.

\section*{Acknowledgement}
We appreciate and thank all the anonymous reviewers for their constructive feedback towards improving this work. The work was
supported in part by the National Language Translation Mission (NLTM): Bhashini project by the Government of India.
\bibliography{custom}

\appendix

\section{Appendix}
\subsection{Data Augmentaion}
\label{DA}
In our data augmentation (DA) experiments, we employ the \textit{Rotation} algorithm described \cite{sahin-steedman-2018-data}. This approach rearranges the siblings of headwords within a defined set of relations. This alters a collection of words or configuration data, but it does not modify the dependencies.

\subsection{Permutation Generation}
\label{permutation}
For generating sentence permutations, we randomly rearrange each word in a sentence to generate phrase permutations while maintaining the relationship between the words. The random permutations are generated while preserving the original dependency tree structures and relations between words in each training sentence. In other words, we first generate the dependency trees for the original sentences and randomly permute the linear order of words, ensuring that the newly permuted sentences still respect the same dependency relations between word pairs.

\subsection{Integration of CSSL with another encoder}
\label{modular nature}
The modular nature of CSSL framework allows for seamless integration with any encoder architecture, without necessitating alterations to pretraining decisions. We have shown its effectiveness for the best-performing baseline. We are also showing results with one more baseline (for Sanskrit). Our supplementary results indicate that activating contrastive loss for the G2GTr baseline on the STBC treebank for Sanskrit leads to an approximate 2-point enhancement in performance measured by UAS/LAS.
\begin{table}[ht]
    \centering
    \begin{tabular}{l|l|l|l|l}
    \hline
        ~ & CE & ~ & CSSL & ~ \\ \hline
        ~ & UAS & LAS & UAS & LAS \\ \hline
        G2GTr & 87.16 & 85.68 & 89.05 & 87.05 \\ \hline
    \end{tabular}
    \caption{Contrastive Loss with G2GTr on STBC dataset.}
    \label{tab:g2gtr_cts}
\end{table}

\label{sec:appendix}
\subsection{Treebank Statistics}
Table \ref{tab:stats} provides the detailed statistics for the languages used in the experiments.
\begin{table*}[!htbp]
    \centering
    \begin{tabular}{c|c|c|c|c}
    \hline
        \multicolumn{1}{c|}{Treebank} & \multicolumn{1}{l|}{Language Family}  &\multicolumn{1}{c|}{train} & \multicolumn{1}{c|}{dev} &\multicolumn{1}{c}{test}  \\ \hline 
        Sanskrit-STBC & Indo-Aryan & 2,800 & 1,000 & 300 \\ \hline
        UD-Turkish\_IMST & Turkic & 3,435 & 1,100 & 1,100 \\ \hline
        UD-Gothic\_Proeil & Germanic & 3,387 & 985 & 1,029 \\ \hline
        UD-Telugu\_MTG & Dravidian & 1,051 & 131 & 146 \\ \hline
        UD-Hungarian\_Szeged & Uralic & 910 & 441 & 449 \\ \hline
        UD-Ancient\_Hebrew\_PTNK & Semitic & 730 & 439 & 410 \\ \hline
        UD-Lithuanian\_ALKSNIS & Baltic & 2,341 & 617 & 684 \\ \hline
        UD-Turkish\_PENN & Turkic & 14850 & 622 & 924 \\ \hline
        UD-English\_EWT & Roman & 12,544 & 2,001 & 2,077 \\ \hline
    \end{tabular}
    \caption{Treebank Statistics. The number of sentences in train, dev and test for each language.}
    \label{tab:stats}
\end{table*}

% \begin{table*}[t]
%     \centering
%     \begin{tabular}{l|ll|ll|ll|ll}
%     \hline
%         Language & CE & ~ & DA & ~ & CTS & ~ & CTS+DA & ~ \\ \hline
%         ~ & UAS & LAS & UAS & LAS & UAS & LAS & UAS & LAS \\ \hline
%         Sanskrit & 89.62 & 87.43 & 90.38 & 88.46 & 91.86 & 89.38 & 92.43 & 90.18 \\ \hline
%     \end{tabular}
%     \caption{Performance comparison on the RNGTr model among standard cross entropy (CE), DA (CE+Data
%     Augmentation),CTS (CE+Contrastive) techniques and CTS+DA techniques on Sanskrit STBC dataset.The results in the CTS+DA setting are statistically significant compared to that of DA and CTS as per the t-test with a p-value < 0.01 for the LAS metric.}
%     \label{tab:cts+da}
% \end{table*}
\subsection{Related Work}
Contrastive learning has
been the pinnacle of recent successes in sentence
representation learning. \cite{chen2020simple} proposed SimCLR by refining the idea of contrastive learning with the help of modern image augmentation techniques to
learn robust sets of features. In order to optimize the appropriately designed contrastive loss functions, \cite{gao-etal-2021-simcse, zhang2022pairwise} uses the entailment sentences in NLI as positive pairs, significantly improving upon the prior state-of-the-art results. To this end, a number of methods have been put forth recently in which the augmentations are obtained through back-translation \cite{fang2020cert}, dropout \cite{yan2021consert,gao-etal-2021-simcse}, surrounding context sampling \cite{logeswaran2018efficient, giorgi-etal-2021-declutr}, or perturbations carried out at different semantic-level \cite{wu2020clear, yan2021consert}.

\end{document}